\newcommand{\CLASSINPUTtoptextmargin}{19mm}
\newcommand{\CLASSINPUTbottomtextmargin}{38mm}
\documentclass[conference,a4paper]{IEEEtran}
\usepackage{cite}
\usepackage{amsmath,amssymb,amsfonts}
\usepackage{algorithmic}
\usepackage{graphicx}
\usepackage{textcomp}
\usepackage{xcolor}
\usepackage{cite}
\usepackage{comment}
\usepackage{subcaption}
\usepackage{multirow}
\def\BibTeX{{\rm B\kern-.05em{\sc i\kern-.025em b}\kern-.08em
    T\kern-.1667em\lower.7ex\hbox{E}\kern-.125emX}}

\IEEEoverridecommandlockouts
\IEEEpubid{\makebox[\columnwidth]{978-1-5386-4633-5/18/\$31.00~\copyright{}2018 IEEE \hfill} \hspace{\columnsep}\makebox[\columnwidth]{ }}

\usepackage[mathlines,switch]{lineno}

\makeatletter

\let\old@ps@IEEEtitlepagestyle\ps@IEEEtitlepagestyle
\def\confheader#1{%
    \def\ps@IEEEtitlepagestyle{%
        \old@ps@IEEEtitlepagestyle%
        \def\@oddhead{\strut\hfill#1\hfill\strut}%
        \def\@evenhead{\strut\hfill#1\hfill\strut}%
    }%
    \ps@headings%
}
\makeatother

\confheader{%
        \parbox{18cm}{Accepted to the 2020 IEEE International Conference on Communications, Control, and Computing Technologies for Smart Grids (SmartGridComm)}
}
\begin{document}

\title{Demand-Side Scheduling Based on Multi-Agent Deep
Actor-Critic Learning for Smart Grids\\
\thanks{This research is supported by the National Research Foundation (NRF), Singapore, under Singapore Energy Market Authority (EMA), Energy Resilience, NRF2017EWT-EP003-041, Singapore NRF2015-NRF-ISF001-2277, Singapore NRF National Satellite of Excellence, Design Science and Technology for Secure Critical Infrastructure NSoE DeST-SCI2019-0007, A*STAR-NTU-SUTD Joint Research Grant on Artificial Intelligence for the Future of Manufacturing RGANS1906, Wallenberg AI, Autonomous Systems and Software Program and Nanyang Technological University (WASP/NTU) under grant M4082187 (4080), Singapore Ministry of Education (MOE) Tier 1 (RG16/20), Alibaba Group through Alibaba Innovative Research (AIR) Program and Alibaba-NTU Singapore Joint Research Institute (JRI).

DOI: 10.1109/SmartGridComm47815.2020.9302935}
}

\author{\IEEEauthorblockN{Joash Lee${^1}$, Wenbo Wang${^2}$, Dusit Niyato${^2}$}
\IEEEauthorblockA{\textit{${^1}$Energy Research Institute @ NTU, Nanyang Technological University, Singapore}\\
\textit{${^2}$ School of Computer Science and Engineering, Nanyang Technological University, Singapore}\\
\{l.joash, wbwang, dniyato\}@ntu.edu.sg}
}

\maketitle

\begin{abstract}
We consider the problem of demand-side energy management, where each household is equipped with a smart meter that is able to schedule home appliances online. The goal is to minimize the overall cost under a real-time pricing scheme. While previous works have introduced centralized approaches in which the scheduling algorithm has full observability, we propose the formulation of a smart grid environment as a Markov game. Each household is a decentralized agent with partial observability, which allows scalability and privacy-preservation in a realistic setting. The grid operator produces a price signal that varies with the energy demand. We propose an extension to a multi-agent, deep actor-critic algorithm to address partial observability and the perceived non-stationarity of the environment from the agent's viewpoint. This algorithm learns a centralized critic that coordinates training of decentralized agents. Our approach thus uses centralized learning but decentralized execution. Simulation results show that our online deep reinforcement learning method can reduce both the peak-to-average ratio of total energy consumed and the cost of electricity for all households based purely on instantaneous observations and a price signal.
\end{abstract}


\section{Introduction}
\label{sec_intro}
In the past decade, Demand-Side Management (DSM) with price-based demand response has been widely considered to be an efficient method of incentivizing users to collectively achieve better grid welfare \cite{vardakas15}. In a typical DSM scenario, an independent utility operator manipulates the aggregate energy demand by adopting a time-varying pricing scheme that is announced to users on a rolling basis. In response to the pricing scheme, the users voluntarily schedule or shift their appliance loads with the aim of minimizing their individual costs (see the example in~\cite{deng15}). Typical objectives can be to minimize energy consumption, costs, peak load demand ~\cite{deng15,ruelens17,wen15,kim16,mocanu19,mohsenian10,barker12} or maximize the utility of certain actors ~\cite{cao18,reddy11,lu18}.

Under the framework of automatic price-demand scheduling, a variety of problem formulations have been previously proposed.  The DSM problem can be formulated under a centralized or decentralized framework. It can also be categorized under a planning-based framework or an online framework that allow users to react to time-variant features. Examples of centralized planning include casting DSM as an optimization problem ~\cite{mohsenian10,cao18}, while an example  of a centralized online approach is the scheduling problem proposed by~\cite{barker12}. However, a shortcoming of these methods is their inability to account for unknown parameters or stochasticity in the DSM system such as random task arrivals and real-time pricing. An alternative method is to model the system dynamics as a Markov Decision Process (MDP) with partially observability~\cite{hansen18}. Instead of planning on the premise of a known system model, strategies can be learned through reinforcement learning~\cite{ruelens17,wen15}.

In this paper, we propose a Markov game-based framework that allows household energy users within a microgrid to manage their energy consumption both decentrally and online via their smart meters. Our second contribution is an extension of a deep actor-critic algorithm to train these smart meter-enabled household agents to minimize the cost of energy under real-time pricing. (We shall also refer to households as 'agents' in the reinforcement learning context from hereon.) Our proposed approach offers two main advantages. The main advantage of decentrally-controlled energy-managing smart meters is that, after the training phase, each household agent is able to act independently in the execution phase, without requiring specific information about other households --- each household agent will not be able to observe the number or nature of devices operated by other households, the amount of energy they individually consume, and when they consume this energy. This allows for privacy-preservation and scalability in the execution phase. The main advantage of the online strategy-learning framework is that each household agent can respond appropriately to fluctuations brought about by stochastic processes within the microgrid system. We introduce details of the system model in Section~\ref{sec_system_model}.

Our aforementioned aim addresses two key challenges. Firstly, the model of the microgrid system and its pricing methodology is not available to each household. Secondly, the limitation in the observation available to each household and the lack of visibility regarding the strategy of other households creates the local impression of a non-stationary environment.  

To overcome the first mentioned key challenge, we propose a model-free policy-based reinforcement learning method in Section~\ref{sec_rl_method} for each household to learn their optimal DSM strategies. To address the second key challenge, we propose a centralized critic network which provides information to guide the learning processes of the households. Experiments are presented in Section~\ref{sec_experiments}. By comparing our method to an offline method that plans consumption based on a predicted price schedule \cite{mohsenian10}, we show that our method is better able to handle the uncertainties associated with real-time pricing.

\section{Related Work and Preliminaries}
A number of previous works have applied reinforcement learning to train decision-making entities in smart grid settings \cite{lu18, kara12, kim16, reddy11}. These studies applied tabular value-based reinforcement learning algorithms to optimize energy costs. Consequently, these methods were limited to processing only discretized state and action spaces.

Recently, there have also been studies that have introduced methods utilizing deep neural networks for multi-agent reinforcement learning, albeit to different classes of problems \cite{RN242,RN243,RN244}. These newly introduced methods are extensions of the now-canonical deep Q-network (DQN) algorithm for the multi-agent environment: they utilize a central value-function approximator to share information. However, as explained in section \ref{sec_rl_method}, a value-based reinforcement learning algorithm would not suit the problem we introduce in this paper because we consider a reward function that cannot be fully described by only the current state and action.

In this paper, we propose a policy-based algorithm based on the Proximal Policy Optimization (PPO) algorithm \cite{RN136} for a multi-agent framework. We consider the partially observable Markov game framework \cite{Littman}, which is an extension of the Markov Decision Process (MDP) for multi-agent systems. A Markov game for $N$ agents can be mathematically described by a tuple $\langle\mathcal{N}, \mathcal{S}, \{\mathcal{A}_n\}_{n\in\mathcal{N}}, \{\mathcal{O}_n\}_{n\in\mathcal{N}}, \mathcal{T}, \{r_n\}_{n\in\mathcal{N}} \rangle$, where $\mathcal{N}$ is the set of individual agents, $n$ is the agent index, and $\mathcal{S}$ describes set of all possible overall states of all agents in the environment.

At each discrete time step $k$, each agent $n$ is able to make an observation $o_n\in\mathcal{O}_n$ of its own state. Based on this observation, each agent takes an action that is chosen using a stochastic policy $\pi_{\theta_n}: \mathcal{O}_n \mapsto \mathcal{A}_n$, where $\mathcal{A}_n$ is the set of all possible actions for agent index $n$.  The environment then determines each agent's reward as a function of the state and each agent's action, such that $r_n(k): \mathcal{S} \times \mathcal{A}_1 \times \cdots \times \mathcal{A}_n \mapsto \mathbb{R} $. It moves on to the next state according to its state transition model $\mathcal{T: S} \times \mathcal{A}_1 \times \cdots \times \mathcal{A}_n \to \mathcal{S}'$.

\section{System Model}
\label{sec_system_model}
We consider a DSM problem in a microgrid of $N$ households that each have $M$ household appliances. Each household is equipped with a smart meter that is capable of controlling all household appliances. The problem that we present is a more advanced form of the demand-side scheduling problem than that presented in~\cite{mohsenian10} where each household has no advanced knowledge of upcoming tasks. We primarily consider interactive background appliances. These appliances are actively turned on by the household, but the actual time of operation poses a less immediate impact to the user. The appliances that we consider in each household can be, for example, washing machines, clothes dryers, storage water heaters, dish washers or refrigerators.

Each household's local operational state can be expressed as $\mathbf{s}_n=[\mathbf{x}_n^{\top},
\mathbf{t}_n^{\top}, \mathbf{l}_n^{\top}, \mathbf{q}_n^{\top}]^{\top}$. It consists of an indicator of whether each household appliance (super-scripted index) is turned on or off $\mathbf{x}_n=[x^1_n,\ldots,x^M_n]^{\top}$, the length of time before the next task can commence $\mathbf{t}_n=[t^1_n,\ldots,t^M_n]^{\top}$, the number of tasks queued for each appliance $\mathbf{q}_n=[q^1_n,\ldots,q^M_n]^{\top}$, and the lengths of the subsequent tasks each appliance has in queue $\mathbf{l}_n=[l^1_n,\ldots,l^M_n]^{\top}$. The overall state is simply the joint operational state of all the users: $\mathbf{s} = [\mathbf{s}_1,\ldots, \mathbf{s}_N]$. We adopt a practical assumption that each household can only observe its own internal state along with the published price at the previous time step $p(k-1)$, such that $\mathbf{o}_n(k) = [\mathbf{s}_n(k)^\top, p(k-1)]^\top$.

The task arrival for each appliance corresponds to the household's demand to turn it on, and is based on a four-week period of actual household data from the Smart* dataset \cite{barker12data}. We first discretise each day into a set of $H$ time intervals $\mathcal{H}=\{0,1,2,...,H-1\}$ of length $T$, where each interval corresponds to the time in the simulated system. The simulation time interval for each time-step in the Markov game is determined with the following relation:
\begin{equation}
  \label{eq_time_map}
  h(k) = \mod (k, H).
\end{equation}

Next, we count the number of events in each interval where the appliances of interest were turned on. This is used to compute an estimate of the probability that each appliance receives a new task during each time interval, $\text{Pr}({q'}_n^m=1|h(k))$. The task arrival for each appliance ${q'}_n^m$ at each time interval $k$ is thus modelled with a Bernoulli distribution. Each task arrival ${q'}_n^m$ for a particular appliance $m$ in household $n$ is added to the queue $q^m_n$.

The duration of the next task to be executed is denoted by $l^m_n$. Its value is updated by randomly sampling from an exponential distribution whenever a new task is generated (i.e. ${q'}_n^m = 1$). The rate parameter $\lambda_n^m$ for the distribution for each appliance is assigned to be the reciprocal of the approximate average duration of tasks in the above-mentioned four-week period selected from the Smart* dataset \cite{barker12data}. We choose to keep the task duration constrained to $l \in [T, \infty]$.

The use of the described probabilistic models for generation of task arrivals and task lengths enable us to capture the variation in household energy demand throughout the course of an average day while preserving stochasticity in energy demand at each household.

We consider that each household's scheduler is synchronized to operate with the same discrete time steps of length $T$. At the start of a time step, should there be a task that has newly arrived into an empty queue ($q^m_n = 1$), the naive strategy of inaction would have the task start immediately such that $t_n^m = 0$. This corresponds with the conventional day-to-day scenario where a household appliance turns on immediately when the user demands so. However, in this paper we consider a setting where each household has an intelligent scheduler that is capable of acting by delaying each appliance's starting time by a continuous time period $a_n^m \in [0,T]$. The joint action of all users is thus: $\mathbf{a} = [\mathbf{a}_1,\ldots, \mathbf{a}_N]^{\top}$.

Once an appliance has started executing a task from its queue, $q_n^m$ is shortened accordingly. We consider that the appliance is uninterruptible and operates with a constant power consumption $P_n^m$. If an appliance is in operation for any length of time during a given time step, we consider its operational state to be $x^m_n=1$. Otherwise, $x_n^m=0$.

Let $d^m_n(k)$ denote the length of time that an appliance $m$ is in operation during time step $k$ for household $n$. The energy consumption of a household $E_n(k)$ during time step $k$ is thus:
\begin{equation}
  \label{eq_power_load}
  E_n(k) = \sum_{m=1}^{M}d^m_n(k) P_n^m.
\end{equation}

On the grid side, we consider a dynamic energy price $p(k)$ that is a linear function of the Peak-Average-Ratio \cite{vardakas15,mohsenian10} of the energy consumption in the previous $\kappa$ time steps:
\begin{equation}
  \label{eq_price}
  p(k)=\frac{\kappa T\max\limits_{k-\kappa +1\le k}\sum\limits_{n=1}^{N}E_n(k)}{\sum\limits_{n=1}^{N}E_n(k)}.
\end{equation}

Each household receives a private reward signal that is a weighted sum of a cost objective and a soft constraint:
\begin{equation}
    \label{eq_reward}
    r_{n}(k) = -r_{c,n}(k) + w r_{e,n}(k).
\end{equation}

The first component of the reward function $r_{c,n}$ is the monetary cost incurred by each household at the end of each time step. The negative sign of the term (\ref{eq_cost}) in  (\ref{eq_reward}) encourages the policy to delay energy consumption until the price is sufficiently low.
\begin{equation}
    \label{eq_cost}
    r_{c,n}(k) = p(k) \times E_n(k)
\end{equation}

The second component of the reward is a soft constraint $r_{e,n}$ tunable by weight $w$. The soft constraint encourages the policy to schedule household appliances at a rate that matches the average rate of task arrival into the queue.
\begin{equation}
    \label{eq_constraint}
    r_{e,n}(k) = E_n(k)
\end{equation}

The state transition for all households can be considered to be Markovian because the transition $\mathcal{T}$ to the next overall system state is dependent on the current state and the actions of all household agents. In contrast, we recall from (\ref{eq_price}) and (\ref{eq_reward}) that the price of energy, and consequently each agent's reward, is a function of not only the current state and actions, but also of the history of previous states. Our pricing-based energy demand management process is, therefore, a generalization of the Markov game formulation \cite{Littman}.

\section{Semi-Distributed Deep Reinforcement Learning for Task Scheduling}
\label{sec_rl_method}
The DSM microgrid problem that we introduced in the previous section presents various constraints and difficulties in solving it: (i) each household can choose its action based only on its local observation and the last published price, (ii) the reward function is dependent on the states and actions beyond the last time step, (iii) the state and observation spaces are continuous, (iv) the household agents have no communication channels between them, and (v) a model of the environment is not available. Constraint (iv) is related to the general problem of environmental non-stationarity: concurrent training of all agents would cause the environmental dynamics to constantly change from the perspective of a single agent, thus violating Markov assumptions.

Requirements (i) and (ii) and environmental non-stationarity mean that value-based algorithms such as Q-learning are not suitable, because these algorithms depend on the Markov assumption that the state transition and reward functions are dependent only on the state and action of a single agent at the last time step. Furthermore, requirement (v) rules out model-based algorithms. Thus, in this section, we propose an extension of the actor-critic algorithm Proximal Policy Optimization (PPO) \cite{RN136} for the multi-agent setting which we will refer to as Multi-Agent PPO (MAPPO).

Let $\mathbf{\pi}=\{\pi_1, \pi_2, ..., \pi_N\}$ be the set of policies, parameterized by $\mathbf{\theta}=\{\theta_1, \theta_2, ..., \theta_N\}$, for each of the $N$ agents in the microgrid. Each policy $\pi_n(\mathbf{a}_n|\mathbf{s}_n)$ is stochastic, in that it produces and samples from a probability distribution of actions given the observed state. The objective function that we seek to maximize is the expected sum of discounted reward  $J(\theta) = E_{\mathbf{s}\sim Pr^\pi,\mathbf{a}\sim \pi_\theta}[\sum^K_{k=i}\gamma^{k-i}r_k]$, where $Pr^\pi$ is the probability distribution of states under policy set $\pi$, and $r_k$ is obtained based on (\ref{eq_reward}). In canonical policy gradient algorithms such as A2C, gradient ascent is performed to improve the policy:
\begin{equation}
    \label{eq_policy_grad}
    \begin{split}
    & \nabla_{\theta_n} J_n(\theta_n)= \\
    & E_{\mathbf{s}\sim Pr^{\pi}, \mathbf{a}_n\sim \pi_n}
    \left[\nabla_{\theta_n} \log\pi_n(\mathbf{a}_n|\mathbf{s}_n) Q^{\pi_n}_n(\mathbf{s}_n, \mathbf{a}_n) \right].
    \end{split}
\end{equation}

This policy gradient estimate is known to have high variance. It has been shown that this problem is exacerbated in multi-agent settings \cite{RN242}. For this reason, we propose the implementation of PPO as it limits the size of each policy update. Additionally, we augment the individually trained actor networks with a central critic network $\hat{V}^\pi$ that approximates the value function for each household $V^\pi_n$ based on the overall system state, thus receiving information for the training of cooperative actions by individual households. The objective function we propose is:
\begin{equation}
    \label{eq_obj_}
    \begin{split}
    L(\theta_n)=E_k
    \Big[ & \min \Big( \rho_k(\theta_n)\hat{A}^\pi_n(k),\max \big( 1-\epsilon, \\ & \min(\rho_k(\theta_n),1+\epsilon) \big) \hat{A}^\pi_n(k)  \Big) +\lambda H({\pi_n}) \Big],
    \end{split}
\end{equation}
where $\rho_k(\theta_n) = \frac{\pi_n( \mathbf{a}_k|\mathbf{o}_k)}{\pi_{n,old(\mathbf{a}_k|\mathbf{o}_k)}}$, and both $\epsilon$ and $\lambda$ are hyperparameters. $\hat{A}^\pi_n(k)$ is the estimated advantage value, which is computed using the critic network:
\begin{equation}
    \label{eq_adv}
    \hat{A}^\pi_n(\mathbf{s}, \mathbf{a}_n) \approx r(\mathbf{s}, \mathbf{a}_n) + \gamma \hat{V}^\pi(\mathbf{s}(k+1))_n - \hat{V}^\pi(\mathbf{s}(k))_n.
\end{equation}

We fit the critic network, parameterized by $\phi$, with the following loss function using gradient descent:
\begin{equation}
    \label{eq_critic_obj}
    L(\phi) = {\sum_{n} \sum_{k} \left( \hat{V}^\pi( \mathbf{s}(k))_n -\mathbf{y}_n(k) \right)^2}.
\end{equation}
where the target value $\mathbf{y}(k)=r_n(k) + \gamma \hat{V}^\pi (\mathbf{s}(k+1))_n$. As the notation suggests, this update is based on the assumption that the policies of the other agents, and hence their values, remain the same. In theory, the target value must be re-calculated every time the critic network is updated. However, in practice, we take a few gradient steps at each iteration of the algorithm.

\begin{figure}[t]
    \centering
    \includegraphics[width=0.45\textwidth]{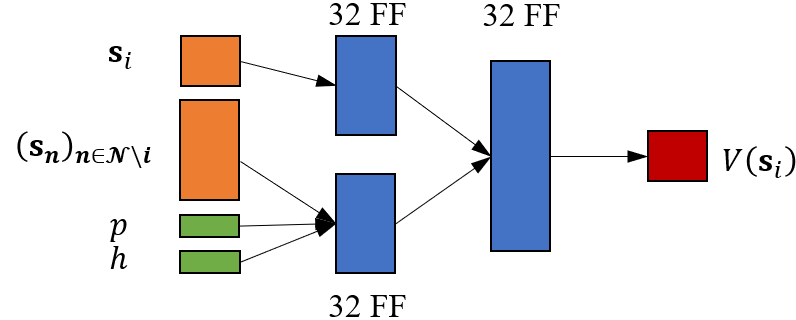}
    \caption{The architecture of a centralized critic neural network.}
    \label{centralized_critic}
\end{figure}

The centralized critic-network utilizes a branched neural network architecture shown in Figure \ref{centralized_critic}. To compute the estimated state value of a particular household $i$, $V_i$, we provide the following input vector to the centralized critic-network:
$[\begin{smallmatrix}
\mathbf{s}_i^\top & \mathbf{s}_1^\top & \mathbf{s}_2^\top & \dots & \mathbf{s}_1^\top & \mathbf{s}_{i-1}^\top & \mathbf{s}_{i+1}^\top & \dots \mathbf{s}_N^\top
\end{smallmatrix}]^\top$. The critic network would thus provide an estimate of the quantity $V^\pi(\mathbf{s})_i = \sum_{k} E_{\pi_\theta} [r(\mathbf{s}_i, \mathbf{a}_i) | \mathbf{s}_1, \mathbf{s}_2, \ldots, \mathbf{s}_{i-1}, \mathbf{s}_{i+1}, \ldots, \mathbf{s}_N]$.

\section{Experiments}
\label{sec_experiments}
In this section, we present the settings and environmental parameters that we use for experimentation, evaluate our proposed Multi-Agent PPO (MAPPO) algorithm against non-learning agents and the offline Energy Consumption Scheduler (ECS) method by \cite{mohsenian10}, examine the effect of augmenting local observations in proposed system model, and consider the effects of household agents learning shared policies.

\subsection{System Setup}
We present experiments on a simulated system of identical households, each with 5 household appliances. As mentioned, their parameters relating to power consumption $P^m_n$, task arrival rates $p^m_n(k)$ and lengths of operation $\lambda^m_n$ are chosen to be a loose match to data sampled from the Smart* 2017 dataset \cite{barker12data}, and thus resemble typical household usage patterns.

\begin{table}[]
\centering
\caption{Time-related hyper-parameters for simulations of the 32-household microgrid system}
\label{hyperparam}
\begin{tabular}{lllcc}
\cline{1-3}
\textbf{Symbol} & \textbf{Hyper-parameter} & \textbf{Value} & \textbf{} & \textbf{} \\ \cline{1-3}
$T$ & Period of each time step & 0.5 hours & \textbf{} & \textbf{} \\
- & Episode length & 240 time steps &  &  \\
- & \begin{tabular}[c]{@{}l@{}}Time steps per iteration of the \\ training algorithm\end{tabular} & 5040 time steps & \multicolumn{1}{l}{} & \multicolumn{1}{l}{} \\ \cline{1-3}
\end{tabular}
\end{table}

Hyper-parameters relating to the simulation time scales are shown in Table \ref{hyperparam}. Each experiment is run with a number of random seeds; the extent of their effect is represented by shaded areas on the graphs shown.

For this chosen experimental setup, we run two baseline performance benchmarks that were non-learning: (i) Firstly, an agent that assigns delay actions of zero for all time steps. This represents the default condition in where the household occupants dictate the exact time that all appliances turn on. (ii) A second baseline agent randomly samples delay actions from a uniform distribution with range $[0, T]$.

\begin{figure}[t]
    \centering
    \includegraphics[width=0.47\textwidth]{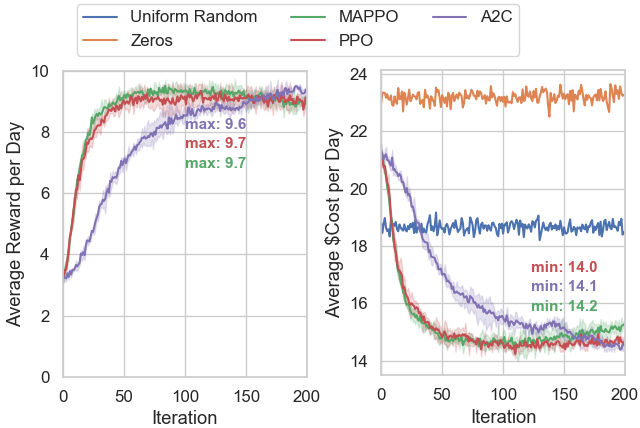}
    \caption{Training progress for a 32-household microgrid system}
    \label{av_cost}
\end{figure}

The results from the training process are compared in Figure~\ref{av_cost}. The y-axes show the  average reward gained (left), and the costs incurred (right) by all households for each day within each batch of samples. All trained policies demonstrate a sharp improvement in reward achieved from the onset of training. The monetary cost decreases with increasing reward, showing that the reward function is suitably chosen for the objective of minimizing cost.

\begin{figure}[t]
\centering     
\includegraphics[width=.45\textwidth]{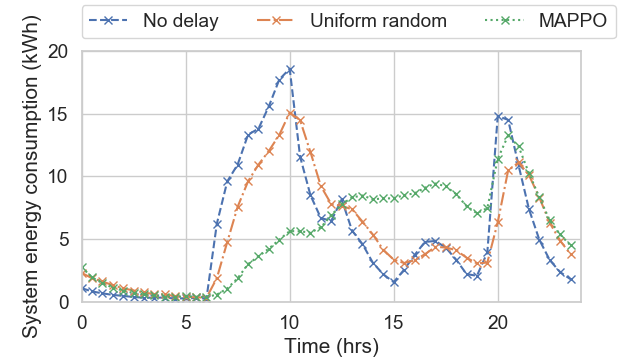}
\caption{Average aggregate energy consumed by 32-household microgrid for each time step of the day.}
\label{av_energy_con}
\end{figure}

In Figure~\ref{av_energy_con}, we plot the average total energy consumed by the entire microgrid system for each time step of the day. Note that the x-axis shows the time in the simulated environment. The blue line shows the energy-time consumption history if all home appliance were to be run immediately on demand; it reveals peaks in demand in the morning and evening, presumably before and after the homes' occupants are out to work, and lows in demand in the night time. The uniform-random-delay agent dampens these peaks in demand by shifting a number of tasks into the following time steps (note the lag in energy consumption compared to the zero-delay agent). In contrast, the agent trained using reinforcement learning is not only able to delay the morning peak demand, but also spread it out across the afternoon. The result is a lower PAR, and hence a lower monetary cost per unit energy.

Although the morning peak in energy demand is smoothed out by the MAPPO-trained households, the same effect is less pronounced for the evening peak. This can be attributed to the training procedure: the batch of experience gathered in each iteration of the algorithm began at 00:00 hrs, and terminated at the end of a 24 hour cycle, providing fewer opportunities for the agents to learn that the evening demand can be delayed to the early hours of the subsequent day. Moreover, delaying tasks on the last cycle of each batch would have led to a lower value of soft-constraint reward $r_{e,n}$, but an un-materialized cost improvement (higher $r_{c,n}$) for the subsequent morning (because of the midnight cut-off).

The results discussed above are significant for our particular application because it shows that while each household is controlled by different policies, all households are able to learn a cooperative strategy that flattens energy demand and consequently decrease the average cost for all users. The ability of the agents to learn in a decentralized manner is compatible with a microgrid system where there may be limited bandwidth to both receive and transmit information centrally. The individual user, or a small cluster of users, could improve their policy using local computing resources, while receiving only the output signal from the centralized critic network. Conversely, if computing resources for neural network training are only available centrally, update gradients can be transmitted to individual actor networks and result in similar performance.

\subsection{Comparison with ECS}
\begin{figure}[t]
\centering     
\includegraphics[width=.45\textwidth]{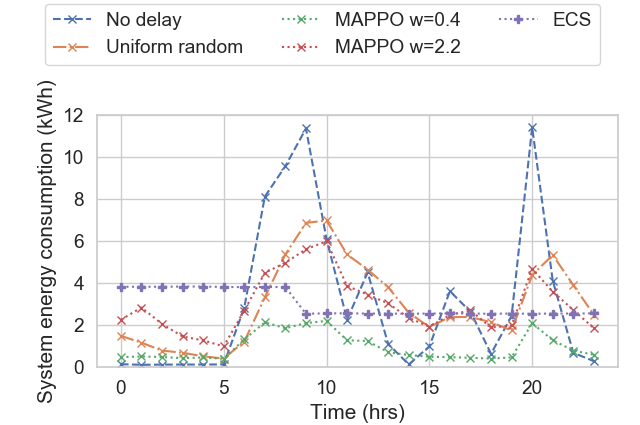}
\caption{Average aggregate energy consumed by 10-household microgrid for each time step of the day}
\label{av_energy_consumed_10houses}
\end{figure}

We compare our proposed MAPPO based method to the Energy Consumption Scheduler (ECS) method proposed by \cite{mohsenian10}. Instead of making decisions online as with MAPPO, ECS requires the following information in advance to schedule all energy consumption for the next day: (i) the mathematical function used to determine the price of electricity for each time step, and (ii) the total planned energy consumption for each household appliance for the entire day, computed as follows:
\begin{equation}
  \label{eq_day_energy}
  \sum_{h=0}^{H-1}E_n^m = \sum_{h=0}^{H-1} \left( \text{Pr}({q'}_n^m=1|h) \times l^m_n \times P^m_n \right).
\end{equation}

For a better comparison, we adapted the experimental setup to match the method used for ECS in its original paper: we considered a system of $N=10$ households, split each day into $H=24$ equal time intervals, fixed the length of operation of each task such that $l^m_n = 1 / \lambda_n^m$, and used a quadratic price function (\ref{eq_price_quad}). We also consider different values of the multiplier $w$ for the cost constraint in the reward function (\ref{eq_reward}).
\begin{equation}
  \label{eq_price_quad}
  p_n(k)=b(h) \times E_n(k)^2.
\end{equation}

In Figure \ref{av_energy_consumed_10houses}, we plot the average aggregate energy consumed by the microgrid for each time step. Once again, the agents trained using reinforcement learning are able to flatten the peaks in energy consumption by spreading demand across subsequent time steps. The MAPPO agents trained with a coefficient of $w=2.2$ consume almost the same energy over the entire day as the naive agents with zero delay; this constitutes a shift in energy demand that results in a decrease in the cost of energy.

\begin{table}[]
\caption{Average reward obtained per day in a 10-household microgrid}
\label{av_reward_10households}
\centering
\begin{tabular}{ccccc}
\hline
\multicolumn{2}{c}{\textbf{Non-learning}} & \multirow{2}{*}{\textbf{\begin{tabular}[c]{@{}c@{}}ECS\\ w=2.2\end{tabular}}} & \multicolumn{2}{c}{\textbf{MAPPO}} \\ \cline{1-2} \cline{4-5} 
\textbf{Uniform Random} & \textbf{Zeros} &  & \textbf{w=0.4} & \textbf{w=2.2} \\ \hline
-0.1 & -13.8 & 7.0 & -0.1 & \underline{8.0} \\ \hline
\multicolumn{1}{l}{} & \multicolumn{1}{l}{} & \multicolumn{1}{l}{} & \multicolumn{1}{l}{} & \multicolumn{1}{l}{}
\end{tabular}
\end{table}

In comparison to the MAPPO agents, ECS produces a flatter energy demand profile. Because the ECS algorithm receives information in advance, it is able to schedule appliances to operate in the early hours of the day to exploit lower prices. However, since the energy consumption inputs received by ECS are average values, each household experiences a mismatch between its actual energy demand and the schedule produced by ECS. Consequently, we compute a reward according to (\ref{eq_reward}), where the value $r_{c,n}$ is taken to be the actual energy demand that is fulfilled by the ECS-generated schedule. The results presented in Table \ref{av_reward_10households} show that the MAPPO agents with $w=2.2$ achieve an average daily reward that is more than $10\%$ higher than that achieved by ECS.

The results reveal the MAPPO method's key advantage, in that it can be utilized in a fully online scenario without requiring the price function or the total daily consumption in advance. It is able to rely on instantaneous observations of the local state and the price signal to determine how tasks should be scheduled by delaying them. This online approach and reliance on mainly locally observable information is more consistent with the nature of domestic electricity use.

Another advantage of MAPPO is that, in addition to shifting the demand of energy, we are able to implement energy reduction by tuning the coefficient $w$. Figure \ref{av_energy_consumed_10houses} shows that the MAPPO agents trained with a smaller coefficient of $w=0.4$ exhibit lower energy demand than both the MAPPO agents with $w=2.2$ and the ECS method.

\subsection{Comparison with Methods using Decentralized Critics}
Figure~\ref{av_cost} also compares our proposed MAPPO algorithm to a similar PPO-based algorithm with decentralized critics, and the more traditional Advantage Actor-Critic (A2C) \cite{pmlr-v48-mniha16} algorithm. The number of learnable parameters across all the $N$ decentralized critic networks was chosen to be similar to that of the centralized critic, on the assumption that this would result in a similar computation resource requirements. The results show that all the agents trained with deep reinforcement learning perform better than the non-learning agents.

Our proposed MAPPO algorithm achieves the most sample-efficient learning and best performance, while the performance of the decentralized PPO algorithm is close. This demonstrates that good performance can be achieved by learning in a completely decentralized manner. This would be useful in scenarios where the connectivity of household smart meters is limited, but local computing power can be utilized.

The A2C algorithm is also able to reach the same level of performance, although it learns less efficiently with the same number of samples. Experiments with higher learning rates showed instabilities that manifested in exploding gradients -- this was in spite of policy-based algorithms such as A2C being generally known to be more stable than value-based algorithms. The reason is that, in our proposed microgrid environment, the updates to the policies for other households under A2C have a large effect on the overall system dynamics, possibly worsening the non-stationarity of the environment as perceived by each household. In contrast, the nature of the actor network update in the PPO algorithm limits the change in the actions taken by the updated policy, thus leading to stable performance increments.

\section{Conclusion}
\label{sec_conclusion}
In this paper, we proposed and implemented a smart grid environment, as well as a multi-agent extension of a deep actor-critic algorithm for training decentralized household agents to schedule household appliances with the aim of minimizing the cost of energy under a real-time pricing scheme. Our approach allows for an online approach to household appliance scheduling that is privacy-preserving, requiring only the local observation and a price signal from the previous time step to act. A joint critic is learned to coordinate training centrally.

Our results show that our proposed algorithm is able to train agents that achieve a lower cost and flatter energy-time profile than non-learning agents. Our algorithm also achieves quicker learning than independent agents trained using the canonical Advantage Actor-Critic (A2C) algorithm. Crucially, its ability to respond to stochastic environmental parameters resulted in increased utility in comparison to an optimization-based planning method.

Future work can involve the extension of the smart grid environment to include more features and real-world complexities; we identify two possibilities. Firstly, the architecture of the microgrid can be expanded to include more energy sources, which may be operated by producers or households. Secondly, we may expand the overall grid to include more agents and more types of agents, such as energy producers and brokers.

\bibliographystyle{IEEEtran}
\bibliography{smart_grid_deep_rl}

\end{document}